\documentclass[10pt,twocolumn,letterpaper]{article}

\usepackage{cvpr}              
\usepackage{multirow}
\usepackage{svg}
\usepackage{tabularx}

\definecolor{cvprblue}{rgb}{0.21,0.49,0.74}
\usepackage[pagebackref,breaklinks,colorlinks,allcolors=cvprblue]{hyperref}
\usepackage{amsmath}

\author{
Miaomiao Cai$^{1}$\footnotemark[1]\quad
Guanjie Wang$^{1}$\footnotemark[1]\quad
Wei Li$^{2}$\footnotemark[2]\quad
Zhijun Tu$^{2}$\quad
Hanting Chen$^{2}$\quad
Shaohui Lin$^{3}$\quad
Jie Hu$^{2}$\footnotemark[2]\\
$^1$University of Science and Technology of China\\
$^2$Huawei Noah’s Ark Lab\\
$^3$East China Normal University\\
\tt\small mmcai@mail.ustc.edu.cn, wei.lee@huawei.com, hujie23@huawei.com}
\def\paperID{8222} 
\def\confName{CVPR}
\def\confYear{2025}

\title{Autoregressive Image Generation with Vision Full-view Prompt}

\begin{document}

\maketitle
\renewcommand{\thefootnote}{\fnsymbol{footnote}}
\footnotetext[1]{Equal Contribution}
\footnotetext[2]{Corresponding Author}
\begin{abstract}
In autoregressive (AR) image generation, models based on the `next-token prediction’ paradigm of LLMs have shown comparable performance to diffusion models by reducing inductive biases. However, directly applying LLMs to complex image generation can struggle with reconstructing the image's structure and details, impacting the generation's accuracy and stability. Additionally, the `next-token prediction’ paradigm in the AR model does not align with the contextual scanning and logical reasoning processes involved in human visual perception, limiting effective image generation. 
Prompt engineering, as a key technique for guiding LLMs, leverages specifically designed prompts to improve model performance on complex natural language processing (NLP) tasks, enhancing accuracy and stability of generation while maintaining contextual coherence and logical consistency, similar to human reasoning. Inspired by prompt engineering from the field of NLP, we propose \textbf
{Vision Full-view prompt} (VF prompt) to enhance autoregressive image generation. Specifically, we design specialized image-related VF prompts for AR image generation to simulate the process of human image creation. This enhances contextual logic ability by allowing the model to first perceive overall distribution information before generating the image, and improve generation stability by increasing the inference steps. Compared to the AR method without VF prompts, our method shows outstanding performance and achieves an approximate improvement of 20\%.
\end{abstract}    
\section{Introduction}
\vspace{-1mm}
Large language models (LLMs) have a significant impact on the field of artificial intelligence, changing approaches to addressing traditional challenges in natural language processing and machine learning~\cite{achiam2023gpt,touvron2023llama}. Recently, in the field of image generation, there has been widespread attention on autoregressive (AR) image generation based on the LLMs.
For example, LlamaGen~\cite{sun2024autoregressive} employs the AR model based on the `next-token prediction' paradigm of large language models for image generation, which reduces the inductive biases on visual signals. Under this design, LlamaGen has demonstrated comparable results to popular diffusion-based image generation models~\cite{peebles2023scalable,rombach2022high}, which shows that the AR model can serve as the foundation for advanced image generation systems.

\begin{figure}[t]
  \centering
   \includegraphics[width=0.9\linewidth]{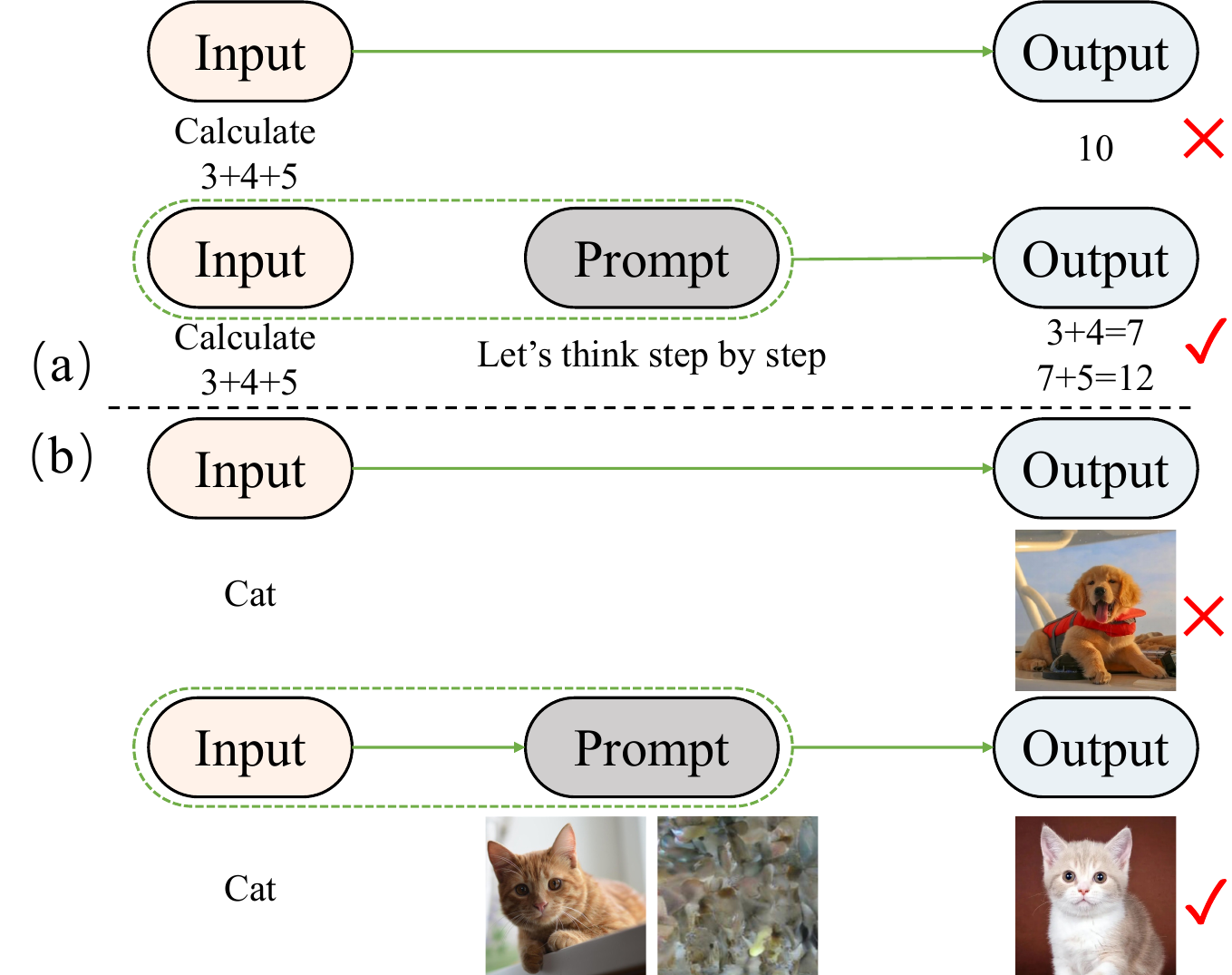}
   \caption{Illustration of the prompt engineering for AR model. The grey module represents prompts. (a) Prompt engineering utilizes designed prompts to perform text tasks. (c) Vision Full-view (VF) Prompt incorporates image-related prompts to perform image generation tasks.}
   \label{teaser}
\end{figure}
\vspace{-1mm} 
However, directly applying LLMs to the image generation task, which is much more complex than the text generation task, could result in the model struggling to reconstruct the structure and details of the image, thus affecting the accuracy of the generated results and the stability of the generation process. 
Additionally, when humans perceive complex visual information, they typically scan the entire image before focusing on the target. In contrast,  the `next-token prediction' paradigm in AR image generation does not align with the contextual sequence and logical reasoning required for real image generation.

In the field of natural language processing (NLP), prompt engineering is a series of techniques that utilize specific designed prompts to enhance the model's generative ability~\cite{sahoo2024systematic,chen2023unleashing,vatsal2024survey}.
As shown in Fig.~\ref{teaser} (a), compared to the general AR model process, prompt engineering leverages specific designed prompts to guide the model, achieving significant performance improvements in complex tasks such as arithmetic, code generation, and knowledge-intensive question answering. 
On the one hand, prompt engineering provides clearer and more informative guidance, reducing objective function drift during training and enhancing training stability. During inference, it reduces uncertainties caused by input, thereby improving the stability of model predictions.
On the other hand, prompt engineering introduces logical guidance and in-context information for input, enabling the model to better simulate human problem-solving approaches. This enhances the coherence of the reasoning process, reduces erroneous reasoning paths, and improves the interpretability and reliability of generated answers.

Despite differing in task formulation from the logical reasoning tasks commonly encountered in NLP-based LLMs, AR image generation can be attributed to a similar organized structure, where each subtask is solved based on the state of the previous one. 
Therefore, inspired by prompt engineering from the field of NLP, to handle complex image generation tasks, we propose a Vision Full-view (VF) prompt based on the AR model, which guides the model to grasp the overall visual information before progressively genrate local details during generation, as illustrated in Fig.~\ref{teaser} (b). 
%
Specifically, we design specialized VF prompt obtained for AR image generation, simulating the human process of creating images by guiding the model to first perceive vision full-view information (such as the universal distribution of the image dataset or representative full-view visual patterns from a specific category) before generating the image. 
Therefore, the VF prompt enhances the contextual logical capability of the model and improves generation stability by increasing inference steps. Experimental results demonstrate that our method significantly outperforms the baseline AR method without prompts, achieving an approximate performance improvement of 20\%.


\section{Related Work}
\subsection{Autoregressive Visual Generation Methods}
Visual generation has received significant attention in recent years, especially with the development of deep learning architectures and generative models. 
One key trend is the use of autoregressive models, which show significant generality and potential due to their strong connection with NLP. However, there is no mature or well-established community, and further efforts are needed to overcome challenges and fully realize their capabilities.
Existing autoregressive generation methods are commonly divided into two approaches: BERT-style mask autoregressive models and GPT-style autoregressive models. 
Mask Autoregressive methods~\cite{chang2022maskgit,yu2023magvit,li2023mage,li2024autoregressive}, inspired by BERT-style pre-training~\cite{kenton2019bert}, generate images by predicting the random masked tokens. Instead, another kind of autoregressive method~\cite{esser2021taming,yu2021vector,ramesh2021zero}, inspired by GPT~\cite{radford2018improving}, predicts the next token in a sequence, which applies the image tokenization~\cite{kingma2013auto,van2017neural} to transform images to discrete space. Recently, based on autoregressive methods, LlamaGen~\cite{sun2024autoregressive} adapts large language model architectures like Llama~\cite{touvron2023llama}, to autoregressively predict image tokens by applying the `next-token prediction' paradigm to visual generation, which achieves decent performance in image generation.

\subsection{Other Visual Generation Methods}
In addition to AR visual generation methods, significant efforts have also been made in exploring other forms of visual generation models.
Generative Adversarial Networks (GANs) are the earliest approaches, leveraging adversarial training to generate images~\cite{goodfellow2014generative,brock2018large,kang2023scaling,sauer2022stylegan,karras2019style}.
Diffusion models, an alternative approach, generates images by gradually refining random noise through a series of learned steps~\cite{song2019generative,song2020denoising,dhariwal2021diffusion,ho2022cascaded,rombach2022high,peebles2023scalable}. GANs-based and diffusion-based methods show promising performance because their community are relatively complete. If AR models are to surpass them, further iterations and development of AR models are needed.

\subsection{Prompt Engineering for LLMs}
Prompt engineering, a critical methodology for optimizing interactions with large language models (LLMs), strategically designs structured instructions to guide models toward high-quality outputs. Previous research has extensively explored various prompting strategies. For example, Zero-shot prompting~\cite{radford2019language} guides the model to complete a task solely through task descriptions, relying on its pre-trained knowledge for reasoning and generation. Few-shot prompting~\cite{brown2020language} provides a small number of examples in addition to the task description, helping the model learn task patterns and improving its performance on specific tasks. Chain-of-Thought (CoT) prompting~\cite{wei2022chain,lyu2023faithful, ye2024diffusion} enhances performance in complex tasks such as mathematics and logical reasoning by instructing the model to generate a step-by-step reasoning process (\textit{e.g.}, `Let's think step by step'). In-context Learning (ICL) prompting~\cite{min2022rethinkingroledemonstrationsmakes,dong2024surveyincontextlearning,pan2023incontextlearninglearnsincontext} enables LLMs to learn task patterns directly from the provided context during inference by including task examples or relevant information in the input, eliminating the need for model parameter updates.
Retrieval-Augmented Generation (RAG)~\cite{lewis2020retrieval,yao2023react} combines information retrieval systems to dynamically fetch external knowledge during inference. This approach allows the model to generate more factually accurate responses and reduces hallucinations.

The reasons why prompt engineering achieves excellent success in the LLM are as follows. First, it refines input instructions to establish a more structured and consistent learning signal, thereby mitigating shifts in the objective function during training and improving overall model stability. Additionally, during inference, it helps reduce variability caused by input fluctuations, thereby improving the stability of model predictions.
Second, prompt engineering introduces logical guidance and in-context information directly into the input enabling the model to align more closely with human-like problem-solving strategies, improving the fluency and coherence of generated responses while minimizing logical inconsistencies and hallucinations.
Based on these two reasons, we apply prompt engineering to autoregressive (AR) image generation to address issues of training instability and inconsistency with human perception in AR image generation.

\section{Method}

\subsection{Preliminary}
\paragraph{AutoRegressive Modeling (AR).} First, an image $I$ in the form of $H\times W \times 3$ is quantized into discrete tokens map $X$ in the form of $h\times w$ by an image tokenizer with $h=H/p, w=W/p$, where p is the down-sampling rate of the tokenizer. Then, according to the raster scanning order, $X$ reshape into 1d sequence $(x_1,x_2,...,x_t), t=h*w$ and the approximate maximum log-likelihood estimation \ref{MLE_vanilla} is used as the training target of the model $\theta$.
\begin{equation}
    \theta_{target} = \arg\max_{\theta} \sum^{T}_{t=1}P_{\theta}(x_t|x_{<t})
    \label{MLE_vanilla}
\end{equation}
In the image generation process, the AR predicts the image token $(x_1,x_2,...,x_t)$ according to the condition $c$ in a prediction manner of `next-token prediction' $\prod^{T}_{t=1}P(x_t|x_{<t},c)$, and finally converts the image token into an image by using a decoder of the image tokenizer.
\subsection{Ours}
\subsubsection{Overview}
Our method still adapts the image tokenizer architecture and `next-token prediction' prediction form of the classic AR model. On this basis, we introduce a set of VF prompts $\Tilde{S}=(s_1,s_2,...,s_k)$ in the training and inference process, and modify the model $\theta$ training target to maximize the log-likelihood function \ref{mle_S}. The way model inference is change to $\prod^{T}_{t=1}P(x_t|x_{<t},\Tilde{S},c)$.
\begin{equation}
    \sum^N_{t=1}\log P_{\theta}(x_t|x_{<t},\Tilde{S},c)
    \label{mle_S}
\end{equation}
\subsubsection{Theoretical Analysis}
\label{theory}
The introduced VF prompt is $\Tilde{S}=(s_1,s_2,...,s_k)$, the generated token sqeuence is $\Tilde{X}=(x_1,x_2,...,x_t)$, the condition is $c$, and the model parameter is $\theta$.

\begin{figure}[t]
  \centering
    \includegraphics[width=\linewidth]{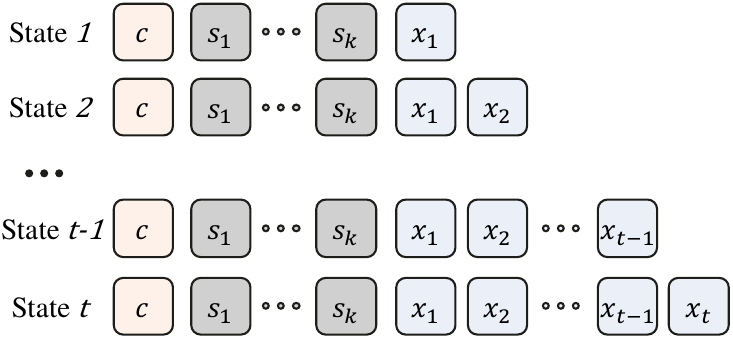}
   \caption{The exhibition of different states with k-length Full-view prompt.}
   \label{fig:state}
\end{figure}

\paragraph{Training Stage} We refer to the derivation of prompt engineering for LLMs in the context of general dynamic programming problems and analyse the impact of introducing prompt engineering in our model while training. As shown in Figure \ref{fig:state}, one sequence shows the different states $i$ with unidirectional transitions between states constrained by the time step $i$. Since there is a target sequence present, for each token prediction during model training, there exists an optimal solution token (\textit{i.e.}, the target token at the corresponding position). Thus, we can express the state transition function as follows:
\begin{equation}
    T(i,j)=
    \begin{cases}
        p_j & \text{if } i=j-1 \\
        p_j*T(i,j-1) & \text{if } i<j-1
    \end{cases}
\end{equation}
$p_j$ represents the probability of the model predicting the optimal state $j$ token at time step $j$. $T(i,j)$ denotes the probability of transitioning from the optimal state $i$ to the optimal state $j$, where the optimal state is defined as a state in which all tokens are the optimal tokens for the current position. Thus, we can formulate $T_{AR}(1,t)$ for the classic AR model and $T_{VF prompt}(1,t)$ for our method as follows:
\begin{align}
\small
   T_{AR}(1,t) &= T(1,2)*T(2,t) \\
    T_{VF\_prompt}(1,t) &= T(c,k)*T(1,2)**T(2,t)
\end{align}
where k represents the length of the sequence $\Tilde{S}$.

It is easy to infer that, under the ideal condition with the same target, both $T_{AR}(1,t)$ and $T_{VF\_prompt}(1,t)$ should equal to the expression $\prod^{T}_{t=1}P(x_t|x_{<t},c)$, which corresponds to the maximum likelihood estimation (\textit{MLE}) of the target generated image. However, since the limitation of the model ability, what we can do is to approximate the target distribution through \textit{MLE}, these tree equations $T_{AR}(1,t)$, $T_{VF\_prompt}(1,t)$ and $\prod^{T}_{t=1}P(x_t|x_{<t},c)$ are not exactly equal. In this case, due to the introduction of VF prompt, our method gains an additional term $T(c,k)$ compared to the classic AR method.

The final loss of the model is constrained by the cross-entropy between the target sequence and the generated sequence $\Tilde{X}$. While it’s not involved in the cross-entropy calculation of classic AR models, it‘s included in VF prompt's cross-entropy calculation, formatted as $\sum \log T(c,k)$. This term, computed by the model, serves as a bias varying with the VF prompt to help the model avoid instability caused by mismatch between training data and inference scenarios. It maintains a consistent optimization direction across different targets, making it faster to converge to an ideal performance.




\begin{figure}[t]
  \centering

\includegraphics[width=\linewidth]{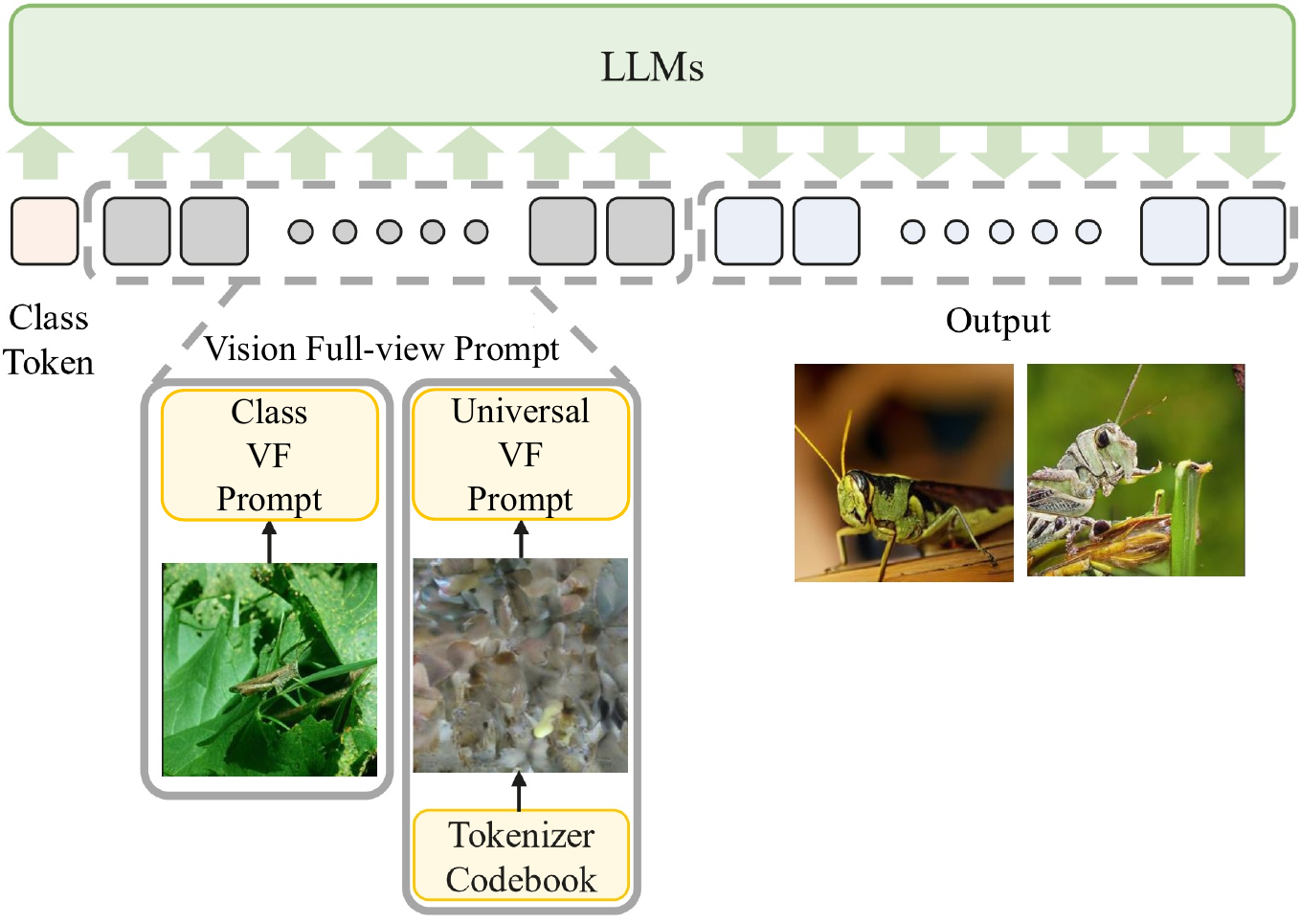}
   \caption{Our method first receives \textit{Class Token} and \textit{Vision Full-view prompt} as input, and then follows the `next-token prediction' to generate images.
}
   \label{fig:architecture}
\end{figure}
\paragraph{Inference Stage} We analyse the impact of introducing VF prompt on the model's inference phase from the perspective of information entropy. The information entropy of the classic AR model during inference can be expressed as $H(\Tilde{X}|c)$. In contrast, the information entropy of our method with VF prompt during inference can be expressed as $H(\Tilde{X}|\Tilde{S},c)$.
With the definition of entropy, we can derive Eq \ref{eq:entropy1} as follows:
\begin{equation}
    H(\Tilde{X}|\Tilde{S},c) = H(\Tilde{X},\Tilde{S}|c) - H(\Tilde{S}|c)
    \label{eq:entropy1}
\end{equation}
Then we can derive the inequality in eq \ref{ieq:entropy2}.
\begin{align}
    &H(\Tilde{X}|\Tilde{S},c) - H(\Tilde{X}|c) = \nonumber \\
    &H(\Tilde{X},\Tilde{S}|c) - H(\Tilde{X}|c) - H(\Tilde{S}|c) \leq 0
    \label{ieq:entropy2}
\end{align}

This inequality suggests that, by incorporating VF prompt, our method has less uncertainty during inference compared to classic AR model, which means the generation process is more stable. This reduction in uncertainty helps effectively minimize inconsistencies in the image details represented by the tokens, leading to higher-quality generated images. The more stable token generation process ensures that the generated sequence better preserves the coherence of the visual structure, resulting in images with fewer artifacts and better alignment with the target distribution.

\subsubsection{Vision Full-view Prompt} 
\label{Prompt}
As shown in Figure \ref{fig:architecture}, we choose two types of vision full-view prompt $\Tilde{S}$ to incorporate into the model's training and inference process.
\begin{enumerate}
    \item We randomly select an image different from the training target while it's still under the same class with the target. After passing it through the image tokenizer, it is converted into tokens, which are then concatenated with the condition $c$ and input into the model. During inference, we randomly choose an image, tokenize it, and use the resulting tokens as input directly. 
    \item We sample a set of indices from the codebook of the image tokenizer based on a uniform distribution, selecting each index with equal probability. The sampled set of indices is then concatenated with the condition $c$ and used as the input to the model.
\end{enumerate}
These two different vision full-view prompts represent two distinct representations used to guide the model in generating images. The first prompt corresponds to a specific category representation called \textit{Class VF prompt}, aimed at helping the model better understand the distribution corresponding to the given condition, thereby generating images that more clearly represent the category information. The second prompt, called \textit{Universal VF prompt}, represents a more universal image distribution and is designed to encourage the model to generate more diverse and varied images, independent of the condition.

\begin{table}
  \centering
  \begin{tabular}{ccccc}
    \toprule
    Model & Parameters & Layers & Hidden Size & Heads \\
    \midrule
    Ours-B & 111M & 12 & 768 & 12 \\
    Ours-L & 343M & 24 & 1024 & 16  \\
    Ours-XL & 775M & 36 & 1280 & 20  \\
    Ours-XXL & 1.4B & 48 & 1536 & 24 \\
    \bottomrule
  \end{tabular}
  \caption{Model sizes and architecture configuration of our method.}
  \label{tab:config}
\end{table}

\begin{table*}[!t]
\centering
\begin{tabular}{>{\centering\arraybackslash}p{4.2cm} >{\centering\arraybackslash}p{4.2cm} >{\centering\arraybackslash}p{1.5cm} >{\centering\arraybackslash}p{1.5cm} >{\centering\arraybackslash}p{1.5cm} >{\centering\arraybackslash}p{1.5cm} }
\toprule
\multicolumn{1}{c}{Training} & Inference & FID$\downarrow$ & IS$\uparrow$ & Precision$\uparrow$ & Recall$\uparrow$ \\ \hline
\multicolumn{2}{c}{Non VF prompt}  & 5.46 & 193.61 & 0.83 & 0.45 \\ \hline
\multirow{2}{*}{Class VF prompt}       & Class VF prompt   &\textbf{4.34} &226.31 &0.84 &0.46 \\ \cline{2-6} 
& Universal VF prompt &  4.36 &227.44 &0.84 &0.46 \\ \hline
\multirow{2}{*}{Universal VF prompt}   & Class VF prompt     & 4.42&217.66&0.85  &0.45  \\ \cline{2-6} 
 & Universal VF prompt &4.39 &222.80 &0.85&0.46 \\ \hline
 \multicolumn{2}{c}{Mixture}&4.55&235.13&0.86&0.44 \\  \bottomrule
\end{tabular}
\caption{Ablation study for different types of VF prompts used for 
 training or inference. `$\downarrow$' or `$\uparrow$' indicate lower or higher values are better. `Mixture' means we combine two different prompts for training and inference,}
\label{tab:Robustness}
\end{table*}
\section{Experiment}
In this section, we describe the implementation details of our approach in Sec.~\ref{sec:implementation}. We then provide ablation studies on important design decisions in Sec.~\ref{sec:Ablation}. Following that, the main results and the visualizations are discussed in Sec.~\ref{sec:MainResults}.
\subsection{Experiment Setting}
\label{sec:implementation}
\paragraph{Model Architecture} Following prior work which uses a VQ tokenizer to tokenize the input images into discrete tokens, we use the VQGAN reported in LlamaGen~\cite{sun2024autoregressive} with the official weight trained on ImageNet. Our model architecture is largely based on Llama~\cite{touvron2023llama}, applying pre-normalization using RMSNorm~\cite{zhang2019rootmeansquarelayer}, SwiGLU~\cite{shazeer2020gluvariantsimprovetransformer} activation function, and 2D rotary positional embeddings~\cite{su2023roformerenhancedtransformerrotary} at each layer of our model. We use models of different configurations including B (111M), L (343M), XL (775M), and XXL (1.4B). The detailed architecture configuration and size are shown in Table~\ref{tab:config}. 

\paragraph{Training Setup} We experiment on ImageNet \cite{ILSVRC15} at a resolution of 256$\times$256 based on the class to image task. To accelerate training, we pre-tokenize the entire training dataset before training using the VQGAN tokenizer reported in LlamaGen~\cite{sun2024autoregressive}. Additionally, we improve the diversity of the pre-tokenized dataset by applying the ten-crop transformation~\cite{sun2024autoregressive}.
Following the classic evaluation suite provided by \cite{dhariwal2021diffusionmodelsbeatgans}, we evaluate FID \cite{heusel2018ganstrainedtimescaleupdate} as the main metric and also report IS \cite{salimans2016improvedtechniquestraininggans}, Precision, and Recall as references metrics.
Unless otherwise specified, models are trained for 300 epochs and the default inference setting is top-k=0, top-p=1.0, temperature=1.0, number of VF prompt=256. Additionally, the dropout for the class condition embedding in Classifier-Free Guidance (CFG) is set to 1.75 by default. All models are trained using the base learning rate of 0.0001 and the batch size of 256. AdamW optimizer is set with $\beta_{1}=0.9$ and $\beta_{2}=0.95$, weight decay is set to 0.05, and gradient clipping is set at 1.0. A dropout rate of 0.1 is applied to the input token embeddings, attention modules, and the FFN module. 

\begin{figure}[t]
  \centering
   \includegraphics[width=\linewidth]{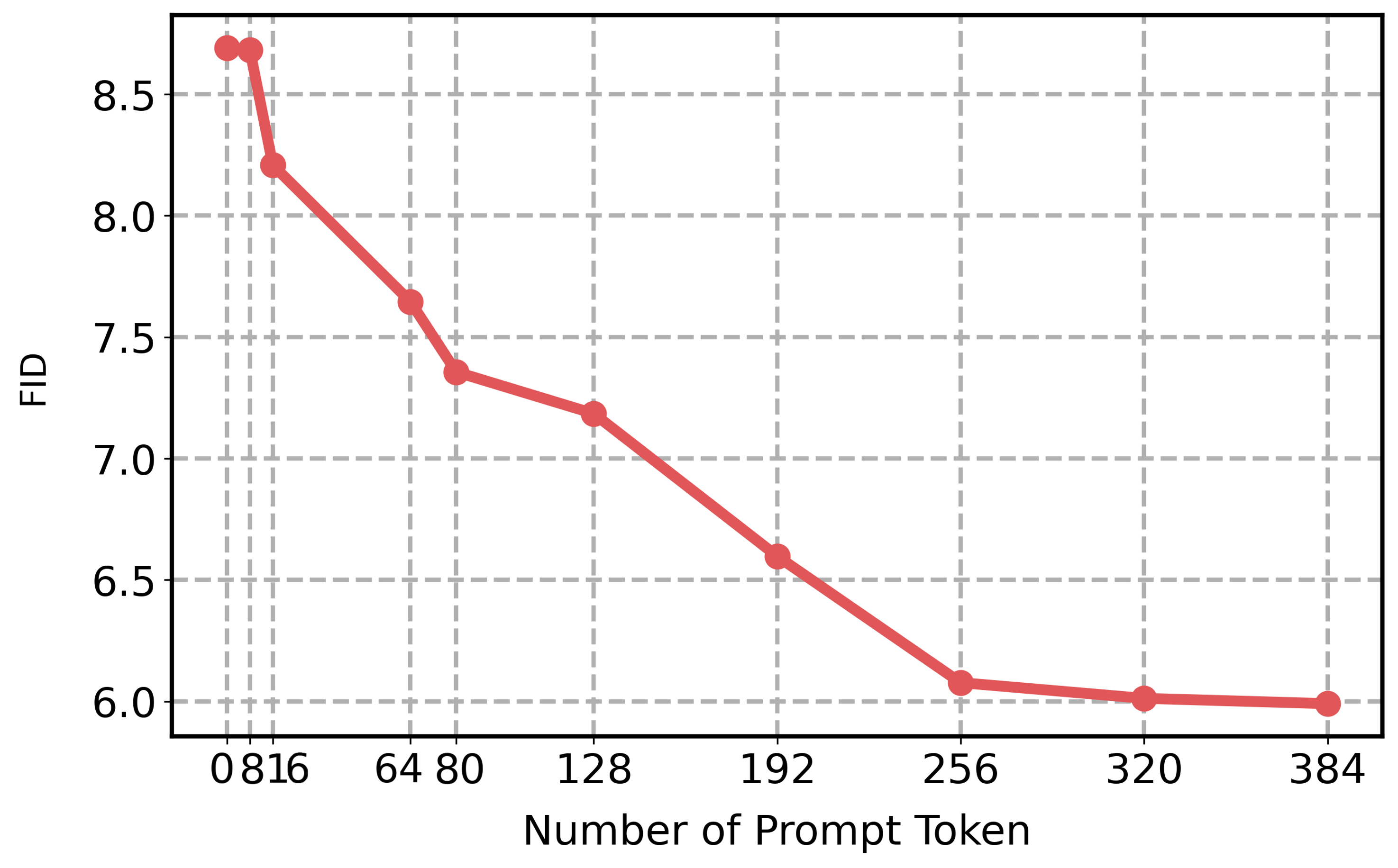}

   \caption{Ablation study for different numbers of prompt tokens.}
   \label{fig:number}
\end{figure}

\subsection{Ablation Study}
This section mainly includes the properties of our method, as well as the choice of parameters.
\label{sec:Ablation}
\subsubsection{Robustness of VF Prompt}
To evaluate the robustness of the proposed VF prompt, we conduct experiments using different types of VF prompts. Specifically, we use two types of prompts mentioned in Sec.~\ref{Prompt}: class VF prompt and universal VF prompt. We evaluate the effectiveness and robustness of VF prompt by comparing the results with consistent VF prompts (\textit{i.e.}, the same VF prompt used during both training and inference) and inconsistent VF prompts (\textit{i.e.}, different VF prompts used during training and inference). 
In addition, we also combine two different prompts for training and inference, with each having a 50\% probability of being selected. We train the model with 111M parameters for 300 epochs, which is the same setting as those in LlamaGen~\cite{sun2024autoregressive}. The baseline is LlamaGen~\cite{sun2024autoregressive}, an AR image generation model based on the Llama architecture without our VF prompt.

\begin{table*}[t]
\centering
\begin{tabular}{>{\centering\arraybackslash}p{4cm}  >{\centering\arraybackslash}p{2cm} >{\centering\arraybackslash}p{2cm} >{\centering\arraybackslash}p{2cm} >{\centering\arraybackslash}p{2cm}}
\toprule
Inference & FID$\downarrow$ & IS$\uparrow$ & Precision$\uparrow$ & Recall$\uparrow$ \\ \hline
Non VF prompt&8.69 &124.44&0.79 &0.47\\
Blank Prompt&8.65&123.15&0.80&0.43\\
Full-view After Generation&8.70&122.79&0.79&0.47 \\
VF prompt&\textbf{6.07}&174.48&0.82&0.44 \\ 
\bottomrule
\end{tabular}
\caption{Ablation study for different variations of prompting. `$\downarrow$' or `$\uparrow$' indicate lower or higher values are better.}
\label{tab:Ablation}
\end{table*}

\begin{table}
  \centering
  \begin{tabularx}{\linewidth}{>{\centering\arraybackslash}X >{\centering\arraybackslash}X >{\centering\arraybackslash}X >{\centering\arraybackslash}X >{\centering\arraybackslash}X}
    \toprule
CFG & FID$\downarrow$ & IS$\uparrow$ & Precision$\uparrow$ & Recall$\uparrow$  \\
    \midrule
    1.25&7.14&106.97
    &0.71
    &0.58\\
    1.50 &5.47&167.61&0.80&0.52   \\
    1.75 &\textbf{4.39} &222.80 &0.85&0.46 \\
    2.00 &5.24&266.85&0.88&0.40 \\
    2.25 &6.83&304.32&0.91&0.36 \\
    2.50&8.49&329.75&0.92&0.31 \\
    \bottomrule
  \end{tabularx}
  \caption{Ablation study for different CFG settings.}
  \label{tab:cfg}
\end{table}

The results are shown in Table~\ref{tab:Robustness}, with the following observations:
(1) Regardless of whether we use consistent or inconsistent VF prompts for training and inference, and regardless of which VF prompt we choose, our method consistently outperforms the baseline without VF prompt, achieving an approximate 20\% improvement in FID scores. This demonstrates that our method exhibits good robustness. 
(2) The performance of the class VF prompt is better than that of the universal VF prompt. This is because focusing on a specific class is more informative than considering the overall distribution of the entire image set, as it more accurately reflects the representation of the specific class. However, the universal VF prompt is more generalizable as it is independent of the condition.
(3) When the VF prompt is consistent between training and inference, the performance is better than when they are inconsistent. This is expected, as consistency between training and inference ensures that the model can better align the reasoning process across both phases, leading to more coherent and stable results.
(4) The inferior performance of the mixed prompt is due to the introduction of information conflict and inconsistency. When using two different prompts, the model needs to balance both, which could lead to confusion in the generated content. Compared to a single prompt, the mixed prompt increases the complexity of reasoning, and the model may not have sufficient capacity to effectively integrate the information from both prompts, thus affecting performance.

\subsubsection{Different Number of Prompt Tokens}
We conduct an ablation study on the length of the intermediate VF prompt, using the universal VF prompt as the VF prompt, which samples a set of indices from the image tokenizer's codebook based on a uniform distribution.  
To save computational resources, all experiments train the model with 111M parameters for 50 epochs, as 50 epochs are sufficient to reflect the performance of different models. The results are shown in Fig.~\ref{fig:number}. A prompt length of 0 corresponds to the baseline, which represents the baseline without VF prompt. 
We can observe that when the VF prompt is short (8 tokens), the ability of VF prompt to handle complex image generation task is weaker because it contains fewer information, resulting in performance similar to the baseline. 
Furthermore, as the length of the prompt increases, the FID performance improves gradually. 
However, the performance gains are not uncapped, which can be found that when the prompt length reaches near 256, the gains begin to slow down.  This phenomenon is reasonable as the gains from using representative information will reach an upper limit.
Therefore, in subsequent experiments, considering both performance and efficiency, the length of VF prompt is set to 256.

\begin{table*}

\centering
{
\setlength{\tabcolsep}{4mm}{
\begin{tabular}{ccccccc}
\toprule
Type & Model & Param & FID$\downarrow$ & IS$\uparrow$ & Precision$\uparrow$ & Recall$\uparrow$ \\
\midrule
\multirow{3}{*}{GAN} & BigGAN~\cite{brock2018large} & 112M & 6.95 & 224.5 & 0.89 & 0.38 \\
\multirow{3}{*}{ } & GigaGAN~\cite{kang2023scaling} & 569M & 3.45 & 225.5 & 0.84 & 0.61\\
\multirow{3}{*}{ } & StyleGAN-XL~\cite{sauer2022stylegan} & 166M & 2.30 & 265.1 & 0.78 & 0.53\\ 

\midrule
\multirow{4}{*}{Diffusion} & ADM~\cite{dhariwal2021diffusion} & 554M & 10.94 & 101.0 & 0.69 & 0.63 \\
\multirow{4}{*}{ } & CDM~\cite{ho2022cascaded} & - & 4.88 & 158.7 & - & -\\
\multirow{4}{*}{ } & LDM-4~\cite{rombach2022high} & 400M & 3.60 & 247.7 & - & -\\
\multirow{4}{*}{ } & DiT-XL/2~\cite{peebles2023scalable} & 675M & 2.27 & 278.2 & 0.83 & 0.57\\

\midrule
\multirow{3}{*}{Maksed AR} & MaskGIT~\cite{chang2022maskgit} & 227M & 6.18 & 182.1 & 0.80 & 0.51 \\
\multirow{3}{*}{ } & MaskGIT-re~\cite{li2023mage} & 227M & 4.02 & 355.6 & - & -\\
\multirow{3}{*}{ } & MAGE~\cite{li2024autoregressive} & 230M & 6.93 & 195.8 & - & -\\

\midrule
\multirow{15}{*}{AR} & VQGAN~\cite{esser2021taming} & 227M & 18.65 & 80.4 & 0.78 & 0.26 \\
\multirow{13}{*}{ } & VQGAN~\cite{esser2021taming} & 1.4B & 15.76 & 74.3 & - & -\\
\multirow{13}{*}{ } & VQGAN-re~\cite{yu2021vector} & 1.4B & 5.20 & 280.3 & - & -\\
\multirow{13}{*}{ } & ViT-VQGAN~\cite{yu2021vector} & 1.7B & 4.17 & 175.1 & - & -\\
\multirow{13}{*}{ } & ViT-VQGAN-re~\cite{yu2021vector} & 1.7B & 3.04 & 227.4 & - & -\\
 \multirow{13}{*}{ }& RQTran.~\cite{lee2022autoregressive} & 3.8B & 7.55 & 80.4 & 0.78 & 0.26\\
 \multirow{13}{*}{ }&RQTran.-re~\cite{lee2022autoregressive} & 3.8B & 3.80 & 323.7 & - & - \\
\multirow{13}{*}{ } & LlamaGen-B (CFG=2.00)~\cite{sun2024autoregressive} & 111M & 5.46 & 193.61 & 0.83 & 0.45\\
 \multirow{13}{*}{ }& LlamaGen-L (CFG=1.75)~\cite{sun2024autoregressive} & 343M &3.81 &248.28 &0.83 &0.52\\
\multirow{13}{*}{ } & LlamaGen-XL (CFG=1.75)~\cite{sun2024autoregressive} & 775M &3.39 &227.08 &0.81 &0.54\\
\multirow{13}{*}{ } & LlamaGen-XXL (CFG=1.75)~\cite{sun2024autoregressive} & 1.4B  &3.09 &253.61  &0.83 &0.53\\
\cline{2-7}

\multirow{3}{*}{ } & VF prompt-B (CFG=1.75) & 111M &4.39 &222.80 &0.85&0.46\\
\multirow{3}{*}{ } & VF prompt-L (CFG=1.75) & 343M&3.23 &279.07 & 0.84& 0.52\\
\multirow{3}{*}{ } & VF prompt-XL (CFG=1.75) & 775M&2.85 &289.75 &0.83 &0.54\\
\multirow{13}{*}{ } & VF prompt-XXL (CFG=1.75) & 1.4B&2.76 &275.80 &0.82 &0.55 \\
\bottomrule
  \end{tabular}}
}
  \caption{Model comparisions on class-conditional ImageNet $256\times256$ benchmark. Metrics include Frechet inception distance (FID), inception score (IS), precision and recall. `-re' means rejection sampling. `$\downarrow$' or `$\uparrow$' indicate lower or higher values are better. `CFG' means using Classifier-Free Guidance. 
  }
  \label{tab:main_results}
\end{table*}

\begin{figure*}[t]
  \centering
   \includegraphics[width=\linewidth]{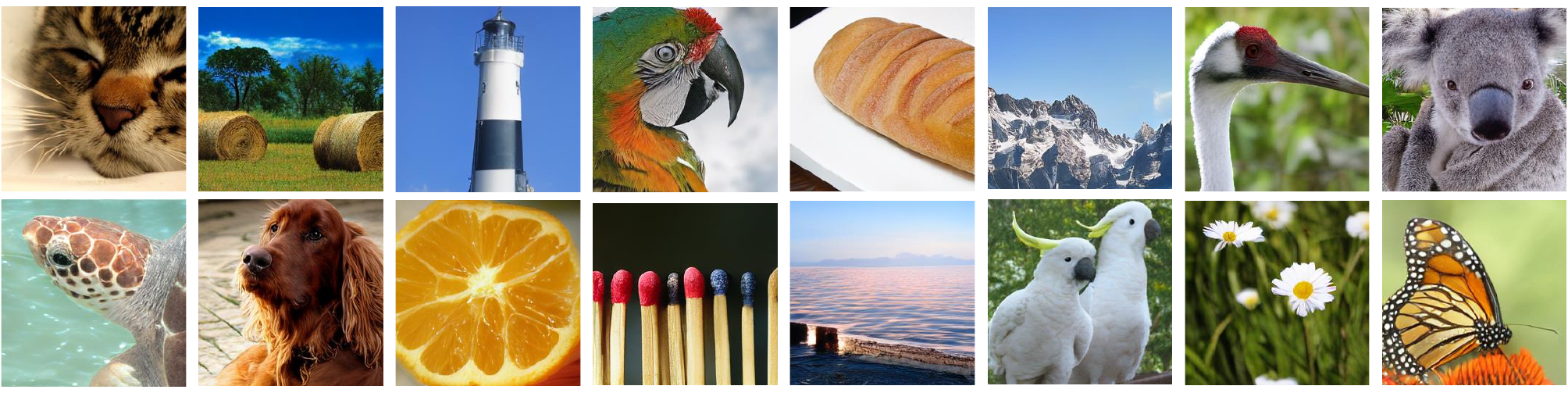}

   \caption{Sample images randomly generated by VF prompt, trained on ImageNet.}
   \label{fig:visual2}
\end{figure*}

\begin{figure}[!t]
  \centering
   \includegraphics[width=0.9\linewidth]{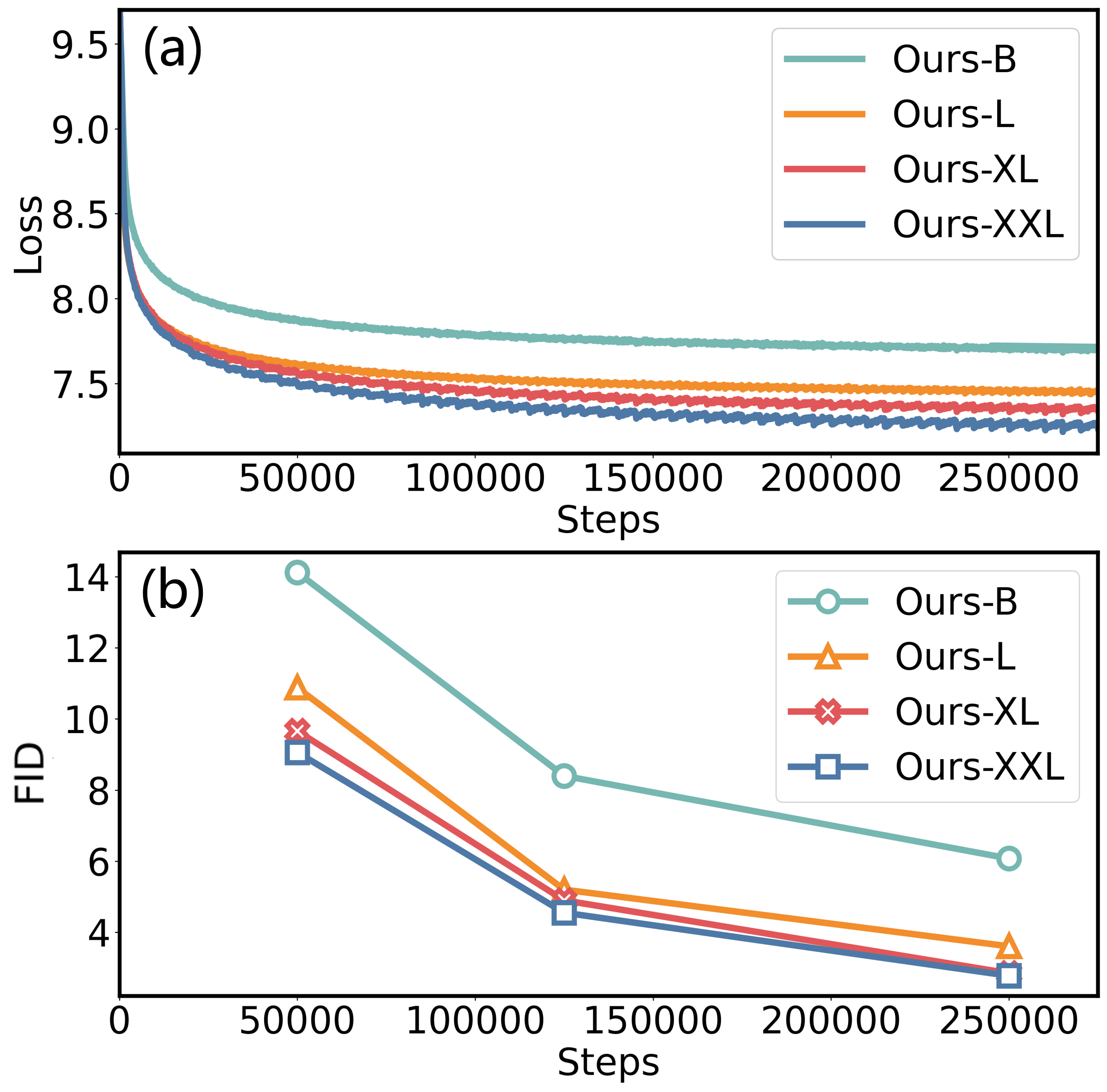}
   \caption{Scaling behavior of VF prompt. (a) Training loss of models with different sizes. (b) FID scores of models with different sizes during inference. 
   }
   \label{fig:scale}
\end{figure}

\begin{figure}[t]
  \centering
   \includegraphics[width=\linewidth]{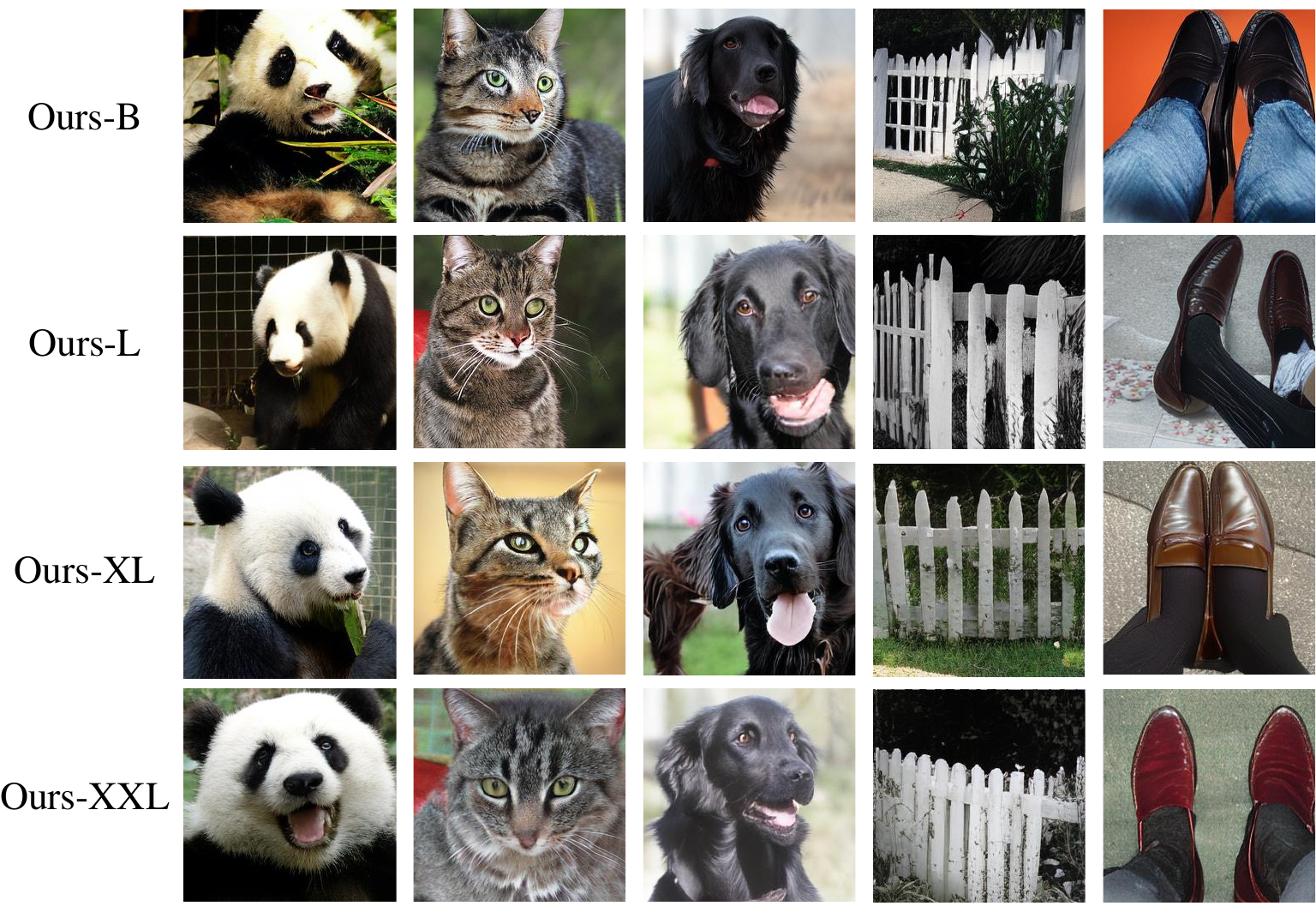}

   \caption{Visualization of samples generated by VF prompt based on different model sizes. 
   }
   \label{fig:visual}
\end{figure}

\subsubsection{Blank Prompt}
Another way to think about our VF prompt is that the proposed image-related VF prompt enables the model to spend more computational resources (\textit{i.e.}, intermediate tokens). To isolate the impact of variable computation from the reasoning process itself, 
we conduct a variant where we replace the VF prompt with multiple blank tokens of equal length, which are tokens extracted from a completely black image with no representation information. The experiments train the model for 50 epochs. As shown in `Blank Prompt' of Table~\ref{tab:Ablation}, the experimental result indicates that this variant performs similarly to the baseline without our VF prompt. 
This demonstrates that variable computation itself is not the reason for the success of VF prompt, and proves the effectiveness of the designed VF prompt, which incorporates full-view image information.

\subsubsection{Order of Full-view and Generation}
A potential benefit of VF prompt is that it allows the model to better access relevant knowledge acquired during pretraining. To investigate this, we test an alternative configuration where the VF prompt is given only after the image generation, isolating whether the model depends on the prior VF prompt to generate the image. The experiments train the model for 50 epochs.
As shown in the `Full-view After Generation' row in Table~\ref{tab:Ablation}, we find that the performance of this variant is comparable to the baseline, suggesting that the sequential full-view itself is valuable in the generation process of the model, rather than merely activating known knowledge. 

\subsubsection{Effect of Classifier-Free Guidance (CFG)}
We conduct ablation experiments under different Classifier-Free Guidance (CFG) settings based on \textit{Universal VF prompt}, as shown in Table~\ref{tab:cfg}. It can be observed that as CFG scale increases, the model's performance gradually improves, with the best FID achieved when CFG scale is set to 1.75. However, further increasing CFG scale leads to a deterioration in FID. Additionally, the increase in CFG scale results in a trade-off between diversity and fidelity, with higher accuracy and lower recall. Therefore, in subsequent experiments, we set the CFG scale to 1.75.

\subsection{Main Results}
\label{sec:MainResults}
\paragraph{Comparisons with Other Image Generation Methods} In Table~\ref{tab:main_results}, we compare with popular image generation models, including GAN~\cite{brock2018large,kang2023scaling,sauer2022stylegan}, Diffusion models~\cite{dhariwal2021diffusion,ho2022cascaded,rombach2022high,peebles2023scalable}, masked AR~\cite{chang2022maskgit,li2023mage,li2024autoregressive} and AR models~\cite{esser2021taming,yu2021vector,lee2022autoregressive,sun2024autoregressive}.
All our experiments are based on the universal VF prompt, as it is more general and easier to obtain compared to the class VF prompt due to its independence from the condition.

The results show that our models outperform all previous AR methods at different levels of model parameters. For example, compared to Llama-B, Llama-L Llama-XL and Llama-XL with the same parameters, VF prompt-B, VF prompt-L VF prompt-XL and VF prompt-XL improves FID by 20\%, 15\%, 16\% and 11\%, and IS by 15\%, 12\%, 28\% and 9\%, respectively. This indicates that our VF prompt of introducing VF prompts effectively helps the AR model in learning and generation by enhancing contextual reasoning of generation and improving stability during the generation process through increased inference steps . 
It is important to note that since LlamaGen does not release the results at 300 epoches for the L and XL models, for a fair comparison, we only present the results for our method at 50 epochs on the XL and XXL models, which have not yet reached the performance limit of our method, leaving room for further improvement.

\paragraph{Visualization} 
As shown in Fig.~\ref{fig:visual2}, we visualize some sample images randomly generated by VF prompt, trained on ImageNet. This demonstrates that VF prompt is capable of generating high-quality samples with both high fidelity and diversity. More visualizations are provided in the appendix.

\paragraph{Scaling Behavior}
We investigate the scaling behavior of VF prompt. As shown in Fig.~\ref{fig:scale}, we show the training loss and FID score based on models of different sizes. It can be observed that VF prompt demonstrates good scalability, with lower training loss and better FID scores as the model size increases. This is because we did not modify the formula or structure of the AR model itself, preserving the integrity of the AR framework, and therefore inheriting the scalability of AR methods. Additionally, we visualize the generation results of the same class under different model sizes, as shown in Fig.~\ref{fig:visual}. It can be observed that as the model size increases, the generated results for the same class become more realistic and detailed, with finer textures and better overall consistency.

\begin{figure}[t]
  \centering
   \includegraphics[width=0.8\linewidth]{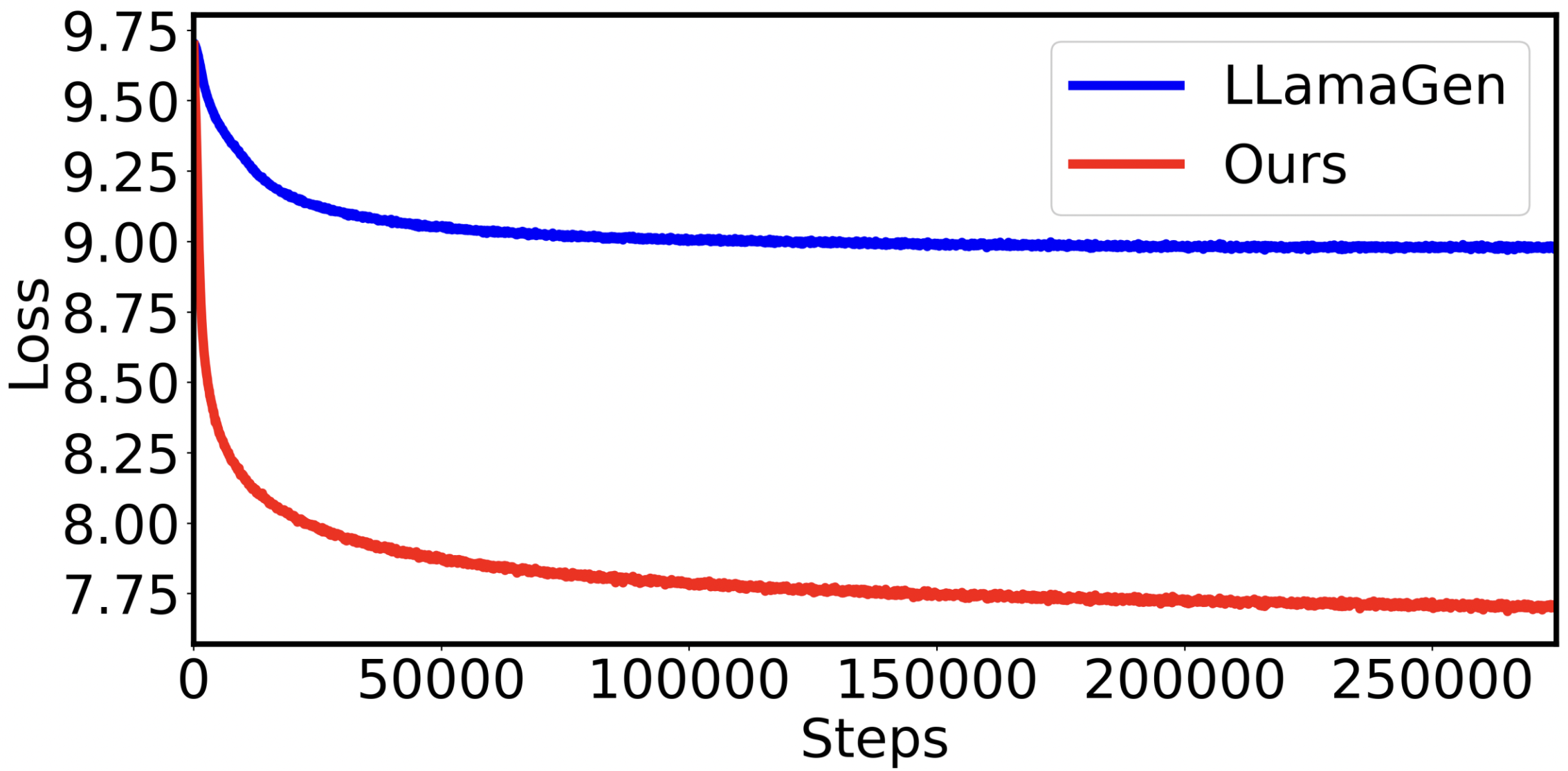}

   \caption{Training loss of the same model with and without VF prompt. 
   }
   \label{fig:loss}
\end{figure}
\paragraph{Convergence}
As the loss curve shown in Fig.~\ref{fig:loss}, it can be observed that the same model architecture after the same training step has a significantly smaller loss for the method using VF prompt, which confirms the conclusions of our theoretical analysis in Sec \ref{theory}.

\section{Conclusion}
In this work, inspired by the prompt engineering from NLP of AR model, we propose the Vision Full-view (VF) prompt, to simulate the human process of creating images by guiding the model to first perceive vision full-view information before generating the image. This approach enhances contextual reasoning by providing a more comprehensive understanding of the overall image information. VF prompt also improves generation stability by increasing the inference steps. VF prompt shows an approximately 20\% improvement over the AR baseline, demonstrating the effectiveness of VF prompts in complex image generation tasks.

{
    \small
    \bibliographystyle{ieeenat_fullname}
    \bibliography{main}
}


\end{document}